\documentclass[runningheads]{llncs}
\usepackage[T1]{fontenc}
\usepackage{graphicx,verbatim}
\usepackage{booktabs} 
\usepackage{multirow} 
\usepackage{pifont}
\usepackage{tabularx}
\usepackage{array}
\usepackage{arydshln}
\usepackage{amsmath}
\usepackage{hyperref}
\usepackage{bm}
\begin{document}
\title{Two-Stage Generative Model for Intracranial Aneurysm Meshes with Morphological Marker Conditioning}
%

\author{
Wenhao Ding\inst{1} \and
Choon Hwai Yap\inst{1} \and
Kangjun Ji\inst{1} \and
Simão Castro\inst{2}
}

\authorrunning{W. Ding et al.} 

\institute{
Imperial College London, London, United Kingdom \\
\email{\{w.ding23, c.yap, kangjun.ji22\}@imperial.ac.uk}
\and
Instituto Superior Técnico, Lisboa, Portugal \\
\email{simao.vitoriano@tecnico.ulisboa.pt}
}

\maketitle              
%
\begin{abstract}
A generative model for the mesh geometry of intracranial aneurysms (IA) is crucial for training networks to predict blood flow forces in real time, which is a key factor affecting disease progression. This need is necessitated by the absence of a large IA image datasets. Existing shape generation methods struggle to capture realistic IA features and ignore the relationship between IA pouches and parent vessels, limiting physiological realism and their generation cannot be controlled to have specific morphological measurements. We propose AneuG, a two-stage Variational Autoencoder (VAE)-based IA mesh generator. In the first stage, AneuG generates low-dimensional Graph Harmonic Deformation (GHD) tokens to encode and reconstruct aneurysm pouch shapes, constrained to morphing energy statistics truths. GHD enables more accurate shape encoding than alternatives. In the second stage, AneuG generates parent vessels conditioned on GHD tokens, by generating vascular centreline and propagating the cross-section. AneuG's IA shape generation can further be conditioned to have specific clinically relevant morphological measurements. This is useful for studies to understand shape variations represented by clinical measurements, and for flow simulation studies to understand effects of specific clinical shape parameters on fluid dynamics. Source code and implementation details are available at \href{https://github.com/anonymousaneug/AneuG}{https://github.com/anonymousaneug/AneuG}.

\end{abstract}

\section{Introduction}
Intracranial aneurysm (IA) is a vascular disease where a weakness in the vascular wall causes a bulge, which carries risks of severe consequences if it ruptures. With advancements in imaging technology, IAs are increasingly detected, with a prevalence of up to 8\% \cite{Lee_2022}. The prediction of rupture risk is currently difficult \cite{Mocco_2018}, but the IA morphological shape and the consequent pattern of blood fluid forces on the IA pouch are thought to influence it \cite{Zhou_2017}. As such, it will be useful to develop neural networks for rapid prediction of flow dynamics in IAs, for use in large clinical data testing of predictive power of fluid dynamics features, and if successful, for subsequent use as a clinical tool to predict rupture risks. To train such a network \cite{Li_2021}, a large dataset of morphologies is necessary, but this is not readily available, necessitating shape generation.

Unfortunately, existing models for synthesizing IA geometries are not yet robust. First, they do not model the joint distribution of IAs and their parent vessel geometries. Instead, they either adopt idealized vessels to merge with generated IA pouches \cite{Goetz_2024} or merge real IAs to healthy parent vessels from other individuals \cite{Nader_2020}. Secondly, some models generate IA pouches via manual deformations \cite{Li_2025}, but this is not data-driven and results may be unrealistic. 

Third, the ability to control the generated IA geometries to achieve specific clinical morphological parameters remains an unmet need. Such controls will allow us to generate shape cohorts with clinically relevant statistics, and understand shape variability within specific clinical measurements. It will also enable the generation of realistic geometries that vary only in specific shape features while keeping all other features constant, which would be invaluable for studies aimed at understanding the effects of individual morphological features on IA fluid mechanics. Current generation models have not achieved this. Past works have been limited to selecting shapes from cohorts generated in an uncontrolled manner \cite{Lv_2020,Qiu_2017,Schnell_2014} or relying on non-physiological synthetic shapes \cite{Goetz_2024,Li_2021_b}.

\begin{figure}
\includegraphics[width=\textwidth]{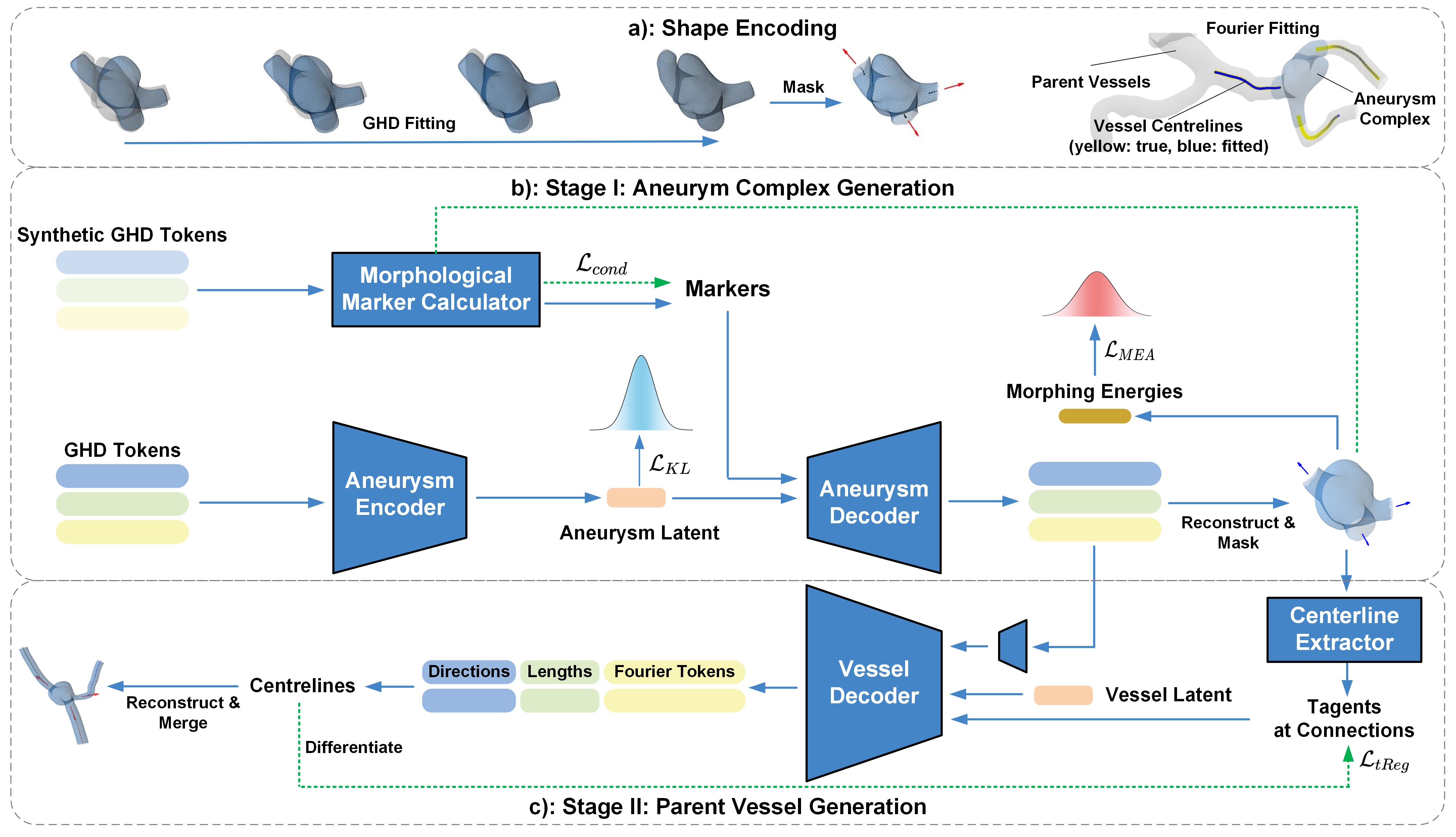}
\caption{Overview of AneuG.} \label{fig1}
\end{figure}

To address these unmet needs, we propose a two-stage conditional generation model, AneuG. In the first stage, the model generates the IA pouch and regions immediately surrounding it — referred to as the aneurysm complex—conditional to having specific values for certain clinically relevant morphological parameters. In the second stage, the parent vessels are generated, conditional to features of the generated aneurysm complex. Both stages employ shape encoding and approximate the latent distributions using a Variational Autoencoder (VAE). This enables accurate reconstruction of 3D shapes with a relatively small dataset, which is challenging for mainstream diffusion-based shape generators \cite{Michelangelo}.

Our main contributions are summarized as follows: (1). We introduce AneuG, the first deep learning generative model for 3D intracranial aneurysm meshes that simultaneously models the aneurysm complex and its parent vessels, where the latter is modelled as the conditional distribution of the former to ensure realistic morphologies. (2). AneuG enables the generation of shapes with specified values of morphological parameters that are clinically significant, via training with a differentiable Morphological Marker Calculator (MMC). This functionality is valuable for downstream morphology and fluid dynamics applications. (3). We utilized the recently developed GHD mesh morphing approach for IA shape encoding, allowing our model to capture more local features of IA shapes. (4). We propose the Morphing Energy Alignment (MEA) constraint to enhance the fidelity to population statistics. This is particularly necessary in conditional generation in (2) where fidelity becomes weaker.

\section{Methods}
\subsection{Problem Statement and Overview}
The schematic of our method is shown in Fig. \ref{fig1}. We aim to generate synthetic aneurysm meshes conditioned on having specific values of certain clinical morphological markers $\bm{\lambda}$, supervised by a cohort of aneurysm shapes reconstructed from MRI. This process is accomplished with a two-stage VAE architecture.

In the first stage, the aneurysm complex undergoes shape feature encoding, in the form of GHD tokens, through a morphing process (see Section \ref{Shape Encoding}). Shapes generated by the unconditional VAE are used to estimate distributions of morphological parameters, which are then used for conditional generation. This ensures a physiologically realistic decoder that understands the conditional probability distributions when multiple parameters are involved. During training, we randomly sample from the latent space and pass the generated shapes through a differentiable Morphological Marker Calculator (MMC) to compute the MSE accuracy of morphology parameters in generated shapes ($\mathcal{L}_{cond}$). Additionally, we calculate the morphing energies of reconstructed shapes and align their distribution with that of real shapes (see Section \ref{Stage I training}).

In the second stage, we generate the centrelines of parent vessels conditioned on the aneurysm complex shape. Similar to stage one, we encode shape features, this time with Fourier tokens. To ensure shape consistency between aneurysm complex and vessels, we calculate the orientations (tangents) of the generated centrelines, and constrain them to match those extracted from the aneurysm complex ($\mathcal L_{tReg}$). This ensures a smoother connection between the aneurysm complex and the generated vessels. Finally, we propagate nodes of the connection cross-sections along the generated centrelines and merge them with the aneurysm complex to obtain the final mesh.


\subsection{Shape Encoding}\label{Shape Encoding}
We encode the 3D aneurysm shapes and their parent vessels before feeding them into VAEs. For the aneurysm complex, we apply the recently published Graph Harmonic Deform (GHD) method \cite{GHD}. A set of scalar tuples $\phi_i=(\phi_{ix},\phi_{iy},\phi_{iz})$, which we refer to as GHD tokens, can be found by morphing a canonical mesh $\mathcal{M}^c$ (defined by nodes $\mathcal{V}_c$ and faces $\mathcal{F}_c$) into a target mesh $\mathcal{M}^{t}$ through gradient descent (see Fig. \ref{fig1} \textbf{a}):
\begin{equation}
    \bm{\phi} = \arg\min_{\bm{\phi}} \mathcal{L}_{GHD} \bigg[ \mathcal{M} \bigg( \mathcal{V} = \sum_{i=1}^n U_i \cdot \phi_i, \mathcal{F}_c \bigg), \mathcal{M}^t \bigg]
\end{equation}
where $\mathcal{V}_c$ and $\mathcal{F}_c$ denote the mesh node coordinates and triangle faces of the canonical shape. $U_i$ is the $i$th eigenvector of the canonical mesh's cotangent graph Laplacian. We truncate the number of eigenvector modes to a small number $n$. $\mathcal{L}_{GHD}$ is the loss function evaluating the distance between the warped canonical mesh and the target mesh. We follow the design in \cite{GHD} and add additional Chamfer Distance constraints on vessel cross-sections to guide the morphing process. Further details are in the GHD citation \cite{GHD}.

For the parent vessels, we encode their centrelines in a similar way. We model the vessel branches as warped beams whose analytical mode shapes are sine functions. 
The centerline function of parent vessels in the global coordinate system is therefore expressed as:
\begin{equation}
    L_k = \mathbf{R_k}(\mathbf{v}_k) \cdot 
    \begin{bmatrix}
    \tilde{x}_k, \sum_i \varphi_i \cdot \psi{ik}, \sum_j \varphi_j \cdot \psi_{jk} \end{bmatrix}^T + \mathbf{c}_k
\end{equation}

\begin{equation}
    \Psi_i(\tilde{x}_k) = \sin(i\pi/l_k *\tilde{x}_k),\ \tilde{x}\in[0, l_k]
\end{equation}
where $k$ represents the branch index and $l_k$ is the unwarped beam length. $\mathbf{v}_k)$ is the normalized direction vector of the branch (from start to end) that parameterizes the rotation matrix $\mathbf{R_k}$ from the local to global coordinate system. The transition $\mathbf{c}_k$ is equal to the coordinate of the vessel cross-section center. An example of a fitted case is visualized in Fig. \ref{fig1} \textbf{a}.

\subsection{Stage I training}\label{Stage I training}

We train a VAE to model the distribution of GHD tokens $\bm{\phi}$. Synthetic aneurysm complex meshes $\mathcal{M}^{\alpha}$ are conditionally reconstructed with $\bm{\phi}$ sampled from the approximated posterior distribution $q_{\theta}^{\alpha}(\bm{\phi}\mid \bm{z}, \bm{\lambda})$:
\begin{equation}
    \mathcal{M}^{\alpha} = \mathcal{M} \bigg( \mathcal{V} = \sum_{i=1}^n U_i \cdot \phi_i, \mathcal{F}_c \bigg)
\end{equation}

The reconstruction loss term consists of the MSE loss of GHD tokens and the Chamfer Distances between the real and reconstructed shapes.

\textbf{Morphing Energy Alignment: }To mitigate the limitations of a small training dataset, we incorporate additional information into the VAE training process. Our hypothesis is that generated shapes, if they closely resemble real shapes, should exhibit a similar population distribution of morphing energies (from canonical to target mesh) as the real shapes. In this work, we utilize rigidity energy $E_{r}$ as defined in \cite{RigidE} and Laplacian smoothness $E_{l}$ as described in \cite{LaplacianE}. Shapiro–Wilk tests conducted on our training dataset reveal that both energy distributions are normal. Based on this observation, we sample from the encoded distribution of real shapes during training and generate corresponding synthetic shapes, compute their morphing energies, and align their distributions with those of the real shapes through KL divergence:
\begin{equation}
    \mathcal{L}_{MEA} = \sum_{i} D_{\text{KL}}\left( \mathcal{N}(\mu_{i}^{\text{real}}, (\sigma_{i}^{\text{real}})^2) \parallel \mathcal{N}(\mu_{i}^{\text{syn}}, (\sigma_{i}^{\text{syn}})^2) \right)
\end{equation}
where $\mu$ and $\sigma ^2$ are population mean and variance of morphing energies ($i\in\{ E_{r}, E_{l}\}$) computed on real / synthetic shapes.

\textbf{Morphological Marker Calculator: }Since GHD is a mesh morphing description of the IA mesh that preserves the number of nodes and mesh connectivity, registration between different aneurysm complexes is easily achieved. This allows us to easily track the neck and dome area of a warped shape and calculate markers in a differentiable way. Apart from neck width (NW) and aspect ratio (AR) \cite{Merritt_2021}, We also introduce the lobulation index (LI) defined as the fraction of dome surface area divided by dome volume (V), as IA lobulation is a clinically consequential feature. In practice, we close the mesh of the aneurysm dome and calculate the volume using the discretized form of Gaussian theorem \cite{GHD}. We use the MMC to approximate the multivariate distribution of the morphological conditions, using synthetic shapes generated with a pre-trained unconditional stage-I VAE. This prevents unphysiological condition combinations, such as the simultaneous occurrence of a large AR and NW, which is rare.

\subsection{Stage II training}\label{Stage II training}
We train another VAE to generate parent vessels. Synthetic centerlines are reconstructed through sampling from the distribution of Fourier mode coefficients $\bm{\varphi}$, the direction vector of the vessel branch $\bm{v}$ (with respect to the tangent vector $t_c$ at the vessel cross-section), and the branch length $l$:
\begin{equation}
    q_{\theta}^{\beta}(\{\bm{\varphi}, \bm{v}, \bm{l} \}\mid \bm{z}, \bm{\phi})
\end{equation}
where the condition is the morphology of aneurysm complex parameterized by $\bm{\phi}$. We then propagate the cross sections of the aneurysm complex along the synthetic centerlines to create tubular meshes and merge them with the masked aneurysm complex mesh to obtain the final mesh. The reconstruction loss consists of MSE loss on tokens and centrelines points. We also conduct spatial differentiation of the reconstructed centrelines and force their tangent directions at the connection to the aneurysm complex to match $t_c$.
\begin{table}
\fontsize{8}{10}\selectfont
\caption{Evaluation of unconditional generation. $\uparrow$ means a larger value is better, and $\downarrow$ otherwise.}\label{tab1}
\begin{tabularx}{\textwidth}{|>{\centering\arraybackslash}p{2cm}|*{6}{>{\centering\arraybackslash}X|}} 
\hline
\textbf{Model} & MEA & FPD$\downarrow$ & KPD$\downarrow$ & TMD$\uparrow$ & $CD_v(10^4)\downarrow$ & $CD_n(10^2)\downarrow$ \\
\hline
\multirow{2}{*}{\textbf{AneuG (ours)}} & \ding{51} & \textbf{6.06} & \textbf{1.10} & 5.89 & \textbf{1.88} & 3.63 \\
                      &  & 8.22 & 2.32 & 4.90 & 1.95 & \textbf{3.53} \\
\hline
\textbf{PCA}                   & -         & 13.52 & 7.78 & \textbf{6.45} & 88.33 & 52.87 \\
\hline
\textbf{Diffusion}             & -         & 45.18     & 62.33     & 25.48    & 10.93      & 124.57     \\
\hline
\end{tabularx}
\end{table}

\section{Experiments}
\textbf{Dataset.} We use the the largest publicly available IA dataset, AneuX, to evaluate our model. As we are focussing on IA located at the middle cerebral artery bifurcation in this work, we extracted 116 IAs within AneurX of this nature for training. Ground truth vascular centrelines are extracted with VMTK.

\textbf{Metrics.} Following \cite{3DS2V}, we adopt Fréchet PointNet++ Distance (FPD) and Kernel PointNet++ Distance (KPD) using a pre-trained PointNet++ to evaluate fidelity of generated latent space distribution. We follow \cite{Zhou_2021} and use Total Mutual Difference (TMD) to evaluate the diversity. For VAE reconstruction evaluation, we use Chamfer Distance on mesh nodes ($CD_v$) and face normals ($CD_n$). For conditioning accuracy, We use the relative L2 error on recalled conditions.

\begin{figure}
\centering
\includegraphics[width=\textwidth]{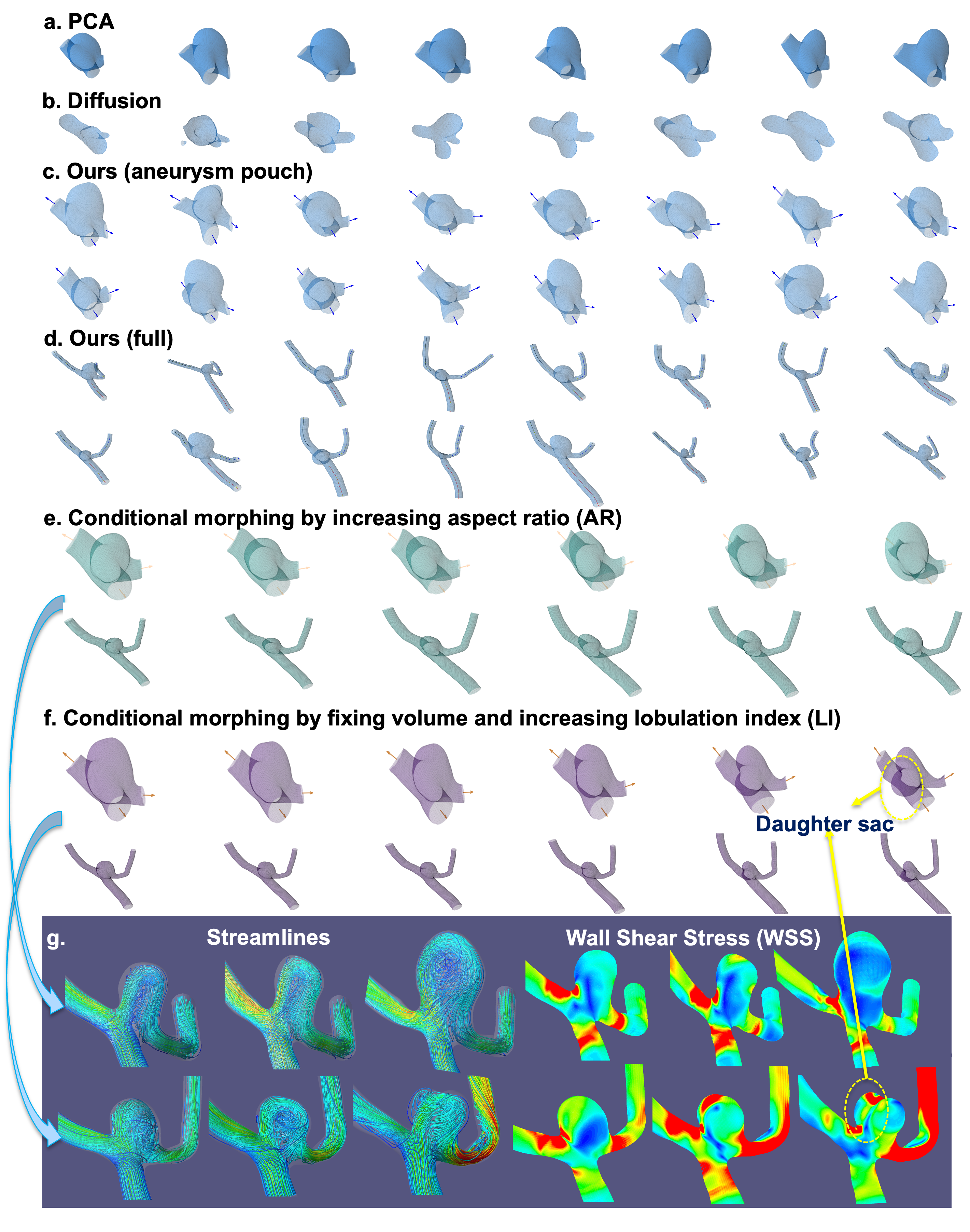}
\caption{Generation gallery. Rotational video for \textbf{c} and \textbf{d} are provided in supplementary material.}\label{fig3}
\end{figure}

\textbf{Unconditional Shape Generation Performance.} Unconditional generation results are presented in Table \ref{tab1}. We compared our AneuG to a generator based on a node-wise Principal Component Analysis (PCA) statistical shape model \cite{Catalano_2022}, which is a prevailing biomedical shape model in the literature, and a deep learning Diffusion shape generator \cite{Michelangelo}. the Diffusion model is composed of a VAE trained on the occupancy field of aneurysm complexes and a transformer UNet-based latent diffusion network. The results indicate that AneuG achieves superior generation fidelity and reconstruction accuracy relative to baseline methods. Results are visualized in Fig. \ref{fig3} \textbf{a}-\textbf{b}, where it is observed that AneuG produces smooth and refined surface meshes, even with only 116 training data, whereas the diffusion model fails to achieve comparable quality. We also observe that PCA achieves smooth generation but fails to capture local features on the aneurysm pouches. These results justify the use of GHD in our approach, which enables AneuG to capture more local features. We further requested a neuroradiologist rank the shape generations from competing models on how realistic they look. AneuG with MEA ranked highest, while AneuG without MEA ranked highest in diversity.

\begin{table}
\fontsize{8}{10}\selectfont
\caption{Evaluation of conditional generation. \textbf{AR}: Aspect ratio. 
\textbf{NW}: Neck width. \textbf{LI}: Lobulation index. \textbf{V}: Dome volume.}\label{tab2}
\begin{tabularx}{\textwidth}{|>{\centering\arraybackslash}p{2cm}|*{12}{>{\centering\arraybackslash}X|}} 
\hline
\textbf{Condition} & \multicolumn{3}{c|}{\textbf{AR}} & \multicolumn{3}{c|}{\textbf{NW}} & \multicolumn{3}{c|}{\textbf{AR \& NW}} & \multicolumn{3}{c|}{\textbf{LI \& V}} \\
\hline
MEA &  & \ding{51} & \ding{51} &  & \ding{51} & \ding{51} &  & \ding{51} & \ding{51} &  & \ding{51} & \ding{51} \\
MMC &  &  & \ding{51} &  &  & \ding{51} &  &  & \ding{51} &  &  & \ding{51} \\
\hline
FPD$\downarrow$ & 14.88 & 9.54 & \textbf{7.86} & 12.41 & 10.50 & \textbf{8.48} & 12.56 & 11.13 & \textbf{10.24} & 17.94 & \textbf{13.20} & 15.16 \\
KPD$\downarrow$ & 11.21 & 4.28 & \textbf{2.78} & 9.27 & 4.66 & \textbf{4.07} & 8.35 & 5.48 & \textbf{4.22} & 21.68 & \textbf{6.79} & 9.22 \\
TMD$\uparrow$ & 3.83 & 4.39 & \textbf{4.56} & 4.10 & 4.43 & \textbf{4.47} & \textbf{4.44} & 4.27 & 4.06 & 3.75 & \textbf{4.01} & 3.13 \\
CA(\%)$\downarrow$ & 8.72 & 9.30 & \textbf{2.93} & 3.35 & 3.68 & \textbf{1.80} & 5.88 & 6.93 & \textbf{2.78} & 11.49 & 12.91 & \textbf{3.19} \\
$CD_v\downarrow$ & 2.79 & \textbf{1.73} & 2.11 & 2.61 & \textbf{1.95} & 3.95 & 2.97 & 2.71 & \textbf{2.06} & 17.18 & 4.68 & \textbf{4.52} \\
$CD_n\downarrow$ & 4.02 & 3.51 & \textbf{3.89} & \textbf{3.45} & 3.48 & 3.69 & 4.40 & 5.12 & \textbf{4.27} & 20.56 & 8.60 & \textbf{8.18} \\
\hline
\end{tabularx}
\end{table}

\textbf{Ablation studies of MEA and MMC.} Conditional generation experiments are conducted to quantify the contributions of MEA and MMC. While the results in Table \ref{tab1} indicate that MEA provides only a marginal improvement in generation fidelity in the unconditional scenario, its impact is more substantial in conditional settings. Table \ref{tab2} demonstrates that incorporating MEA consistently enhances generation fidelity under all condition configurations. This improvement can be attributed to MEA mitigating posterior collapse during conditional training in the first stage, which otherwise reduces the preservation of shape structures and leads to a misaligned morphing energy distribution. By ensuring alignment of the morphing energy distribution with the dataset, MEA helps AneuG maintain more structural details. MMC demonstrates increased conditioning accuracy across all experimental configurations, with a minor improvement in generation fidelity observed in most cases.

\textbf{Application.} By encoding a real shape and fixing the latent code while varying the condition variable, we can morph it to varying morphological parameters. In Fig. \ref{fig3} \textbf{e}, we demonstrate that increasing AR incrementally results in a higher aneurysm dome. In Fig. \ref{fig3} \textbf{f}, we show that by holding the volume of the real shape constant, but increasing lobulation index, a daughter sac can be created, which is a marker of  IA rupture risk \cite{Mocco_2018}. Additionally, Fig. \ref{fig3}\textbf{g} shows the model's ability to generate long vascular inlets and outlets, which is necessary for downstream biofluid mechanics simulation studies as they strongly influence fluid dynamics. We performed flow simulations with the generated shapes, and find that higher IA aspect ratio (AR) causes increased areas of low wall shear stress (WSS), while the presence of a daughter sac causes more complex and chaotic flow. This corroborates prevailing thinking: these flow features are considered mechanical stimuli that can cause adverse biological responses \cite{Zhou_2017}, while these morphology features are understood to be markers for disease progression and rupture risks \cite{Etminan_2015}. These applications highlight how AneuG's conditional morphing abilities can be useful for bridging biomechanics and neurology. In future, large-scale flow simulations can be conducted on controlled shape generations, to further understand how IA biomechanics contribute to rupture and progression risks.

\section{Conclusion}
In this paper, we propose a conditional generation model for intracranial aneurysms. The model captures the joint distribution of aneurysm complexes and their parent vessels, providing a novel solution to this field. The proposed method outperforms the main-stream diffusion model with limited data and therefore highlights its applicability to rare cerebral aneurysm complexes. Future research directions include incorporating temporal conditioning with longitudinal data and leveraging shape knowledge to synthesize images.

%
%
%

\begin{thebibliography}{8}

\bibitem{Lee_2022}
Lee, K.S., Zhang, J.J.Y., Nguyen, V., Han, J., Johnson, J.N., Kirollos, R., Teo, M.: The Evolution of Intracranial Aneurysm Treatment Techniques and Future Directions. Neurosurgical Review \textbf{45}(1), 1--25 (2022).

\bibitem{Mocco_2018}
Mocco, J., Brown, R.D., Torner, J.C., Capuano, A.W., Fargen, K.M., Raghavan, M.L., Piepgras, D.G., Meissner, I., Huston, J.: Aneurysm Morphology and Prediction of Rupture: An International Study of Unruptured Intracranial Aneurysms Analysis. Neurosurgery \textbf{82}(4), 491--496 (2018).

\bibitem{Zhou_2017}
Zhou, G., Zhu, Y., Yin, Y., Su, M., Li, M.: Association of Wall Shear Stress with Intracranial Aneurysm Rupture: Systematic Review and Meta-Analysis. Scientific Reports \textbf{7}(1), 5331 (2017). doi: 10.1038/s41598-017-05886-w. Erratum in: Scientific Reports \textbf{8}(1), 5244 (2018).

\bibitem{Li_2021}
Li, G., Wang, H., Zhang, M. et al.: Prediction of 3D Cardiovascular Hemodynamics Before and After Coronary Artery Bypass Surgery via Deep Learning. Communications Biology \textbf{4}, 99 (2021).

\bibitem{Merritt_2021}
Merritt, W.C., Berns, H.F., Ducruet, A.F., Becker, T.A.: Definitions of intracranial aneurysm size and morphology: A call for standardization. Surgical Neurology International \textbf{12}, 506 (2021).

\bibitem{Goetz_2024}
Goetz, A., Jeken-Rico, P., Pelissier, U., Chau, Y., Sédat, J., Hachem, E.: AnXplore: a comprehensive fluid-structure interaction study of 101 intracranial aneurysms. Frontiers in Bioengineering and Biotechnology \textbf{12} (2024).

\bibitem{Nader_2020}
Nader, R., Autrusseau, F.: Building a Synthetic Vascular Model: Evaluation in an Intracranial Aneurysms Detection Scenario. arXiv preprint arXiv:2411.02477 (2024)

\bibitem{Li_2025}
Li, X., Zhou, Y., Xiao, F., Guo, X., Zhang, Y., Jiang, C., Ge, J., Wang, X., Wang, Q., Zhang, T., Lin, C., Cheng, Y., Qi, Y.: Aneumo: A Large-Scale Comprehensive Synthetic Dataset of Aneurysm Hemodynamics. arXiv preprint arXiv:2501.09980 (2025).

\bibitem{Lv_2020}
Lv, N., Karmonik, C., Chen, S. et al.: Wall Enhancement, Hemodynamics, and Morphology in Unruptured Intracranial Aneurysms with High Rupture Risk. Translational Stroke Research \textbf{11}, 882--889 (2020).

\bibitem{Qiu_2017}
Qiu, T., Jin, G., Xing, H. et al.: Association between hemodynamics, morphology, and rupture risk of intracranial aneurysms: a computational fluid modeling study. Neurological Sciences \textbf{38}, 1009--1018 (2017).

\bibitem{Schnell_2014}
Schnell, S., Ansari, S.A., Vakil, P., Wasielewski, M., Carr, M.L., Hurley, M.C., Bendok, B.R., Batjer, H., Carroll, T.J., Carr, J., Markl, M.: Three-dimensional hemodynamics in intracranial aneurysms: influence of size and morphology. Journal of Magnetic Resonance Imaging \textbf{39}(1), 120--131 (2014).

\bibitem{Li_2021_b}
Li, G., Song, X., Wang, H., Liu, S., Ji, J., Guo, Y., Qiao, A., Liu, Y., Wang, X.: Prediction of Cerebral Aneurysm Hemodynamics With Porous-Medium Models of Flow-Diverting Stents via Deep Learning. Frontiers in Physiology \textbf{12}, 733444 (2021).

\bibitem{3DS2V}
Zhang, B., Tang, J., Niessner, M., Wonka, P.: 3DShape2VecSet: A 3D Shape Representation for Neural Fields and Generative Diffusion Models. arXiv preprint arXiv:2301.11445 (2023).

\bibitem{Catalano_2022}
Catalano, C., Agnese, V., Gentile, G., Raffa, G.M., Pilato, M., Pasta, S.: Atlas-Based Evaluation of Hemodynamic in Ascending Thoracic Aortic Aneurysms. Applied Sciences \textbf{12}(1), 394 (2022).

\bibitem{Michelangelo}
Zhao, Z., Liu, W., Chen, X., Zeng, X., Wang, R., Cheng, P., Fu, B., Chen, T., Yu, G., Gao, S.: Michelangelo: Conditional 3D Shape Generation based on Shape-Image-Text Aligned Latent Representation. In: Advances in Neural Information Processing Systems \textbf{36}, pp. 73969--73982. Curran Associates, Inc. (2023).

\bibitem{GHD}
Luo, Y., Sesija, D., Wang, F., Wu, Y., Ding, W., Huang, J., et al.: Explicit Differentiable Slicing and Global Deformation for Cardiac Mesh Reconstruction. arXiv preprint arXiv:2409.02070v2 (2024).

\bibitem{RigidE}
Sorkine, O., Alexa, M.: As-Rigid-As-Possible Surface Modeling. In: Belyaev, A., Garland, M. (eds.) Geometry Processing, pp. 109--116. The Eurographics Association (2007).

\bibitem{LaplacianE}
Desbrun, M., Meyer, M., Schröder, P., Barr, A.H.: Implicit fairing of irregular meshes using diffusion and curvature flow. In: Proceedings of the 26th Annual Conference on Computer Graphics and Interactive Techniques (SIGGRAPH '99), pp. 317--324. ACM Press/Addison-Wesley Publishing Co. (1999).

\bibitem{GCN}
Kipf, T.N., Welling, M.: Semi-Supervised Classification with Graph Convolutional Networks. arXiv preprint arXiv:1609.02907 (2017). \url{https://arxiv.org/abs/1609.02907}

\bibitem{Zhou_2021}
Zhou, L., Du, Y., Wu, J.: 3D Shape Generation and Completion Through Point-Voxel Diffusion. In: Proceedings of the IEEE/CVF International Conference on Computer Vision (ICCV), pp. 5826--5835. IEEE (2021).

\bibitem{Etminan_2015}
Etminan, N., Brown, R.D., Jr., Beseoglu, K., et al.: The Unruptured Intracranial Aneurysm Treatment Score: A Multidisciplinary Consensus. Neurology \textbf{85}(10), 881--889 (2015).
\end{thebibliography}
%

\end{document}